\documentclass[10pt,conference,a4paper]{IEEEtran}
\IEEEoverridecommandlockouts
\usepackage{cite}
\usepackage{amsmath,amssymb,amsfonts}
\usepackage{algorithmic}
\usepackage{graphicx}
\usepackage{textcomp}
\usepackage{subfigure}
\usepackage{xcolor}
\usepackage{tikz}
\usepackage[compact]{titlesec}
\usepackage{hyperref}
\hypersetup{
    colorlinks=true,
    linkcolor=,
}
\newtheorem{theorem}{Theorem}

\def\etal{\emph{et al}.}
\def\BibTeX{{\rm B\kern-.05em{\sc i\kern-.025em b}\kern-.08em
    T\kern-.1667em\lower.7ex\hbox{E}\kern-.125emX}}
\begin{document}

\title{Reducing the Variance of Variational Estimates of Mutual Information by Limiting the Critic's Hypothesis Space to RKHS}

\author{\IEEEauthorblockN{P Aditya Sreekar,
Ujjwal Tiwari, Anoop Namboodiri}
\IEEEauthorblockA{Center for Visual Information Technology\\
International Institute of Information Technology, Hyderabad\\
Email: paditya.sreekar@research.iiit.ac.in, ujjwal.t@research.iiit.ac.in, anoop@iiit.ac.in}}

\maketitle

\begin{abstract}
Mutual information (MI) is an information-theoretic measure of dependency between two random variables. Several methods to estimate MI from samples of two random variables with unknown underlying probability distributions have been proposed in the literature. Recent methods realize parametric probability distributions or critic as a neural network to approximate unknown density ratios. These approximated density ratios are used to estimate different variational lower bounds of MI. While, these estimation methods are reliable when the true MI is low, they tend to produce high variance estimates when the true MI is high. We argue that the high variance characteristics is due to the uncontrolled complexity of the critic's hypothesis space. In support of this argument, we use the data-driven \textit{Rademacher complexity} of the hypothesis space associated with the critic's architecture to analyse \textit{generalization error bound} of variational lower bound estimates of MI. In the proposed work, we show that it is possible to negate the high variance characteristics of these estimators by constraining the critic's hypothesis space to \textit{Reproducing Hilbert Kernel Space} (RKHS), which corresponds to a kernel learned using \textit{Automated Spectral Kernel Learning} (ASKL). By analysing the generalization error bounds, we augment the overall optimisation objective with effective regularisation term. We empirically demonstrate the efficacy of this regularization in enforcing proper bias variance tradeoff on four different variational lower bounds of MI, namely NWJ, MINE, JS and SMILE. 
\end{abstract}

\section{Introduction}
Mutual information is a fundamental information theoretic measure that quantifies the dependency between two random variables (RVs). Given, two RVs, $X$ and $Y$, mutual information (MI), denoted by $I(X;Y)$ is defined as:

\begin{equation*}
    I(X;Y) = \int_{\mathcal{X\times Y}} log\frac{d\mathbb{P}_{XY}}{d\mathbb{P}_X\otimes\mathbb{P}_Y} d\mathbb{P}_{X,Y}
\end{equation*}
Where, $\mathbb{P}_{XY}$ is the joint probability distribution and, $\mathbb{P}_X$ and $\mathbb{P}_Y$ are the corresponding marginal distributions. Mutual information, $I(X;Y)$ between any two RVs ranges from $0$ to $+\infty$. $I(X; Y)$ is high when $X$ and $Y$ share considerable information or in other words have a high degree of dependency and vice-versa. It is equal to zero \textit{iff} $X$ and $Y$ are mutually independent. MI has found wide applications in representation learning \cite{chen2018isolating,kim2018disentangling,alemi2017fixing,hjelm2018learning}, generative modeling \cite{chen2016infogan}, predictive modeling \cite{Li_2019_CVPR}, and information bottleneck \cite{tishby2000information,tishby2015deep,alemi2016deep}. In the learning from data paradigm, data driven  applications use sample based estimation of MI, where the key challenge is in estimating MI from the samples of two random variables with unknown joint and marginal distributions.

In the big data regime, with continuous increase in sample size and data dimentionality, reliable estimation of MI using mini-batch stochastic optimisation techniques is an area of active research \cite{oord2018representation,belghazi2018mutual,poole2018variational,song2019understanding,ghimire2020reliable}.
Classical non-parametric MI estimators that used methods like binning \cite{fraser1986independent},  kernel density estimation \cite{moon1995estimation} and K-Nearest Neighbour based entropy estimation \cite{kraskov2004estimating} are computationally expensive, produce unreliable estimates, and do not conform to mini-batch based optimisation strategies. To overcome these difficulties, recent estimation methods \cite{belghazi2018mutual,poole2018variational,song2019understanding} couple neural networks with variational lower bounds of MI \cite{nguyen2010estimating,donsker1983asymptotic} for differential and tractable estimation of MI. In these methods, a critic parameterized as a neural network is trained to approximate unknown density ratios. The approximated density ratios are used to estimate different variational lower bounds of MI. Belghazi \etal \cite{belghazi2018mutual}, Poole \etal \cite{poole2018variational} and Song \etal \cite{song2019understanding} consider the universal approximation property of the critic neural network to estimate tighter variational lower bounds of MI. However, universal approximation ability of neural networks comes at the cost of neglecting the effect of critic's unbounded complexity on variational estimation of mutul information, which leads to unstable and highly fluctuating estimates. Similar observations have been reported in literature by Ghimire \etal in \cite{ghimire2020reliable}.  

Nguyen \etal \cite{nguyen2010estimating} by analysing the bias-variance tradeoff of variational lower bound estimates of MI showed the need to regularise the complexity of the critic's hypothesis space for stable and low variance estimation of MI. Motivated by their work, we argue that these variational lower bound estimators exhibit high sensitivity to the complexity of critic's (Neural Network) hypothesis space when optimised using mini-batch stochastic gradient strategy. To support this argument, we use a data-driven measure of hypothesis space complexity called \textit{Rademacher complexity} to bound the generalization error for variational lower bounds of MI. Using these bounds, it is shown that higher complexity of critic's hypothesis space leads to higher generalization error and hence high variance estimates. In this proposal, our critic's hypothesis space is constructed in a smooth family of functions, the \textit{Reproducing Kernel Hilbert Space} (RKHS). This corresponds to learning a kernel using \textit{Automated Spectral Kernel Learning} (ASKL)
\cite{li2019automated}. ASKL parameterized functions in the RKHS as a neural network with cosine activation in the hidden layer. By using the Rademacher complexity of ASKL-RKHS, an effective regularization to control the complexity of the critic's hypothesis space has also been proposed.

Rest of the paper is organised as follows. Related literature has been reviewed in section \ref{related_work}. In section \ref{preliminary}, we explain some crucial concepts related to our work. The discussion on related work and preliminaries is followed by a detailed explanation of our approach  in the section \ref{our_approach} where we present a thorough theoretical analysis. Supporting experimental results are demonstrated in section \ref{experiments}. For the sake of brevity, all proofs related to our proposal are included in the Appendix.   

\section{Related Work}
\label{related_work}
\subsection{Mutual Information Estimation}

Mutual information can be characterized as the KL divergence between joint distribution $\mathbb{P}_{XY}$ and the product of marginal distributions $\mathbb{P}_X\otimes\mathbb{P}_Y$, $I\left(X;Y\right) = D_{KL}\left(\mathbb{P}_{XY}  \middle\|\mathbb{P}_X\otimes\mathbb{P}_Y\right)$. This is the central theme in the derivation of lower bounds of MI from variational lower bounds of KL divergence. KL divergence between two multivariate probability distributions, say $\mathbb{P}$ and $\mathbb{Q}$ belongs to broader class of divergences known as the f-divergences, which are characterized by convex function $f$ of likelihood ratio ($d\mathbb{P}/d\mathbb{Q}$). Nguyen \etal \cite{nguyen2010estimating} formulated variational lower bound of f-divergences by using the convex conjugate of $f$ and leveraged convex empirical optimization to estimate f-divergences. Belghazi \etal \cite{belghazi2018mutual} proposed a tighter variational lower bound of KL divergence which is derived from the Donsker-Varadhan \cite{donsker1983asymptotic} dual representation. In their work, two MI estimators are constructed by optimizing neural network critics to maximize (1) convex conjugate lower bound, and (2) Donsker-Varadhan lower bound. In the proposed work, convex conjugate based lower bound estimator is referred to as NWJ and Donsker-Varadhan based estimator as MINE. Poole \etal{} \cite{poole2018variational} developed a unified framework for different MI estimates and created an interpolated lower bound for better bias-variance tradeoff. They also proposed a lower bound which is optimized using GAN discriminator objective \cite{goodfellow2014generative} to estimate density ratios. We refer to the estimator based on this lower bound as JS. Song \etal{} \cite{song2019understanding} showed that variance of both MINE and NWJ estimators increase exponentially with increase in the true magnitude of MI. The explained cause of this behaviour is the increase in variance of  the partition function estimates \cite{song2019understanding}. They also proposed a lower bound estimator with improved bias-variance tradeoff by clipping the partition function estimate. In the proposed work, we refer to estimator based on this lower bound as SMILE.

In this approach, instead of designing a better lower bound estimate as proposed in \cite{nguyen2010estimating,belghazi2018mutual,poole2018variational,song2019understanding}, we study the effect of restricting the hypothesis space of critics to RKHS for favourable bias-variance tradeoff. The comparative performance of the proposed work reflects the effectiveness of the proposed approach in learning low variance estimates of MI. Similar to this approach, Ghimire \etal \cite{ghimire2020reliable} and Ahuja \etal \cite{ahuja2019estimating} also restricted critic hypothesis space to RKHS. Their methods differ from ours in the choice of kernel functions under consideration. Convex combination of Gaussian kernels were considered in \cite{ahuja2019estimating}. A stationary Gaussian kernel with inputs transformed by a neural network with randomly sampled output weights has been proposed in \cite{ghimire2020reliable}. In contrast to the work, we learn a kernel belonging to a much broader class of non-stationary kernels rather than restricting the kernel to Gaussian kernels.

\subsection{Kernel Learning}
Kernel methods play an important role in machine learning \cite{shawe2004kernel,scholkopf2001learning}. Initial attempts included learning convex \cite{lanckriet2004learning,cortes2012l2} or non linear combination \cite{cortes2009learning} of multiple kernels. While the aforementioned kernel learning methods are an improvement over the isotropic kernels, they cannot be used to adapt any arbitrary stationary kernel. To alleviate this problem \cite{wilson2013gaussian,lazaro2010sparse} proposed approximating kernels by learning a spectral distribution. At the core of these methods is Bochner's theorem  \cite{rudin1962fourier}, which states that there exists a duality between stationary kernels and distributions in spectral domain (Fourier domain). Similarly, Yaglom's theorem \cite{yaglom1987correlation} states that there is a duality between the class of kernels and positive semi-definite functions with finite variations in spectral domain. Kom Samo \etal \cite{samo2015generalized} showed that kernels constructed using Yaglom's theorem are dense in the space of kernels. Ton \etal{} \cite{ton2018spatial} used Monte- Carlo integration and Yaglom's theorem to construct non-stationary kernels for Gaussian Processes. 
Recent methods combine deep learning with kernel learning methods. Deep Kernel Learning \cite{wilson2016deep} placed a plain deep neural network as the front-end of a spectral mixture kernel to extract features, which is further extended to a kernel interpolation framework \cite{wilson2015kernel} and stochastic variational inference \cite{wilson2016stochastic}. Chun-Liang Li \etal{} \cite{li2019implicit} modeled the spectral distribution as an implicit generative model parameterized by a neural network and approximated a stationary kernel by performing Monte-Carlo integration using samples from the implicit model. Hui Xue \etal \cite{xue2019deep} and Jian Li \etal \cite{li2019automated} (ASKL) represented a non-stationary kernel as Monte-Carlo integration of fixed samples which are optimized using gradient descent methods. In this work, ASKL is used to learn the kernel corresponding to the critic's hypothesis space in Reproducing Kernel Hilbert Space.

\section{Preliminary}
\label{preliminary}

\subsection{Variation Lower Bounds of Mutual Information}
\label{prelim_mi}
In this subsection, four different variational lower bounds namely $I_{NWJ}$, $I_{MINE}$, $I_{JS}$ and $I_{SMILE}$ based estimators of MI have been discussed. These estimators are used in throughout this work. In estimating variational lower bounds of MI, a parametric probability distribution or critic $f_\theta$ with trainable parameters $\theta$ is optimised to approximate the likelihood density ratio between the joint and product of marginal distributions ($d\mathbb{P}_{XY}/d\mathbb{P}_X\otimes\mathbb{P}_Y$). The approximated density ratio is used for sample based estimation of MI. The optimisation objective is to maximize the different variational lower bounds of MI with respect to the critic parameters $\theta$ to estimate MI.  

Donsker-Varadhan dual representation \cite{donsker1983asymptotic} based variational lower bound of MI, denoted as $I_{DV}$ is given by: 

\begin{multline}
    \label{eq:IDV}
    I(X;Y) \geq I_{DV}\left(f_\theta\right) = \mathbb{E}_{\mathbb{P}_{XY}}\left[f_\theta(x,y)\right] -\\ log\left(\mathbb{E}_{\mathbb{P}_X\otimes\mathbb{P}_Y}\left[e^{f_\theta(x,y)}\right]\right) 
\end{multline}
The optimal critic for which the equality $I_{DV} = I\left(X;Y\right)$ holds in (\ref{eq:IDV}) is given by  $f_{DV}^*=log\left(d\mathbb{P}_{XY}/d\mathbb{P}_X\otimes\mathbb{P}_Y\right)$. $I_{MINE}$ and $I_{NWJ}$ lower bounds can be derived from Tractable Unnormalized Barber and Argakov (TUBA) lower bound, $I_{TUBA}$, considering only constant positive baseline in \cite{poole2018variational}, that is $a>0$ in the $I_{TUBA}$ formulation defined as: 

\begin{multline}
    \label{eq:ITUBA}
    I(X;Y) \geq I_{TUBA}\left(f_\theta\right) = \mathbb{E}_{\mathbb{P}_{XY}}\left[f_\theta(x,y)\right] -\\ \frac{\mathbb{E}_{\mathbb{P}_X\otimes\mathbb{P}_Y}\left[e^{f_\theta(x,y)}\right]}{a} - log(a) + 1 
\end{multline}

Optimal critic satisfying the equality $I_{TUBA} = I\left(X;Y\right)$ in equation \ref{eq:ITUBA} is given by, $f_{TUBA}^*=log\left(d\mathbb{P}_{XY}/d\mathbb{P}_X\otimes\mathbb{P}_Y\right) + log\left(a\right)$. In this work, $I_{MINE}$ is formulated from $I_{TUBA}$ by fixing the parameter $a$ in (\ref{eq:ITUBA}) as exponential moving average of $e^{f_\theta(x,y)}$ across mini-batches. Similarly, $I_{NWJ}$ is formulated from $I_{TUBA}$ by substituting the parameter $a = e$.

Unlike the methods described above that maximize the variational lower bounds to learn likelihood density ratio, other methods \cite{poole2018variational,song2019understanding,kim2018disentangling} approximate the density ratio for sample based estimation of MI by optimizing GAN discriminator objective defined as:

\begin{multline}
        \label{eq:ganobjective}
    \max\limits_{\theta} \mathbb{E}_{\mathbb{P}_{XY}} \left[log\left(\sigma(f_\theta(x,y))\right)\right] + \\ \mathbb{E}_{\mathbb{P}_X\times \mathbb{P}_Y} \left[log\left(1-\sigma(f_\theta(x,y))\right)\right]
\end{multline}

Where, $\sigma()$ is the sigmoid function. The optimal critic maximizing the GAN discriminator objective is given by, $f_{GAN}^* = log\left(d\mathbb{P}_{XY}/d\mathbb{P}_X\times\mathbb{P}_Y\right)$. Poole \etal \cite{poole2018variational} observed that $f_{NWJ}^*=f_{GAN}^*+1$, where $f_{NWJ}^*$ is the optimal critic for $I_{NWJ}$ and constructed another variational lower bound $I_{JS}$ by substituting $f_{GAN}(x,y)+1$ as the critic function $f_\theta$ into (\ref{eq:ITUBA}). The $f_{GAN}$ is optimized using the GAN discriminator objective. Similarly, Song \etal \cite{song2019understanding} constructed another lower bound of MI, denoted as $I_{SMILE}$ by substituting $f_{GAN}$ as critic $f_\theta$ in $I_{DV}$ expressed in (\ref{eq:IDV}). In \cite{song2019understanding}, the bias-variance tradeoff is controlled by clipping the critic output. It is essential to note that we do not clip the output of the ASKL critic to analyse the effectiveness of restricting the critic function $f_\theta$ hypothesis space to Reproducing Kernel Hilbert Space in controlling bias-variance tradeoff. 

\subsection{Automated Spectral Kernel Learning}
\label{prelim_rkhs}
\begin{figure}
    \centering
    \includegraphics[width=0.47\textwidth]{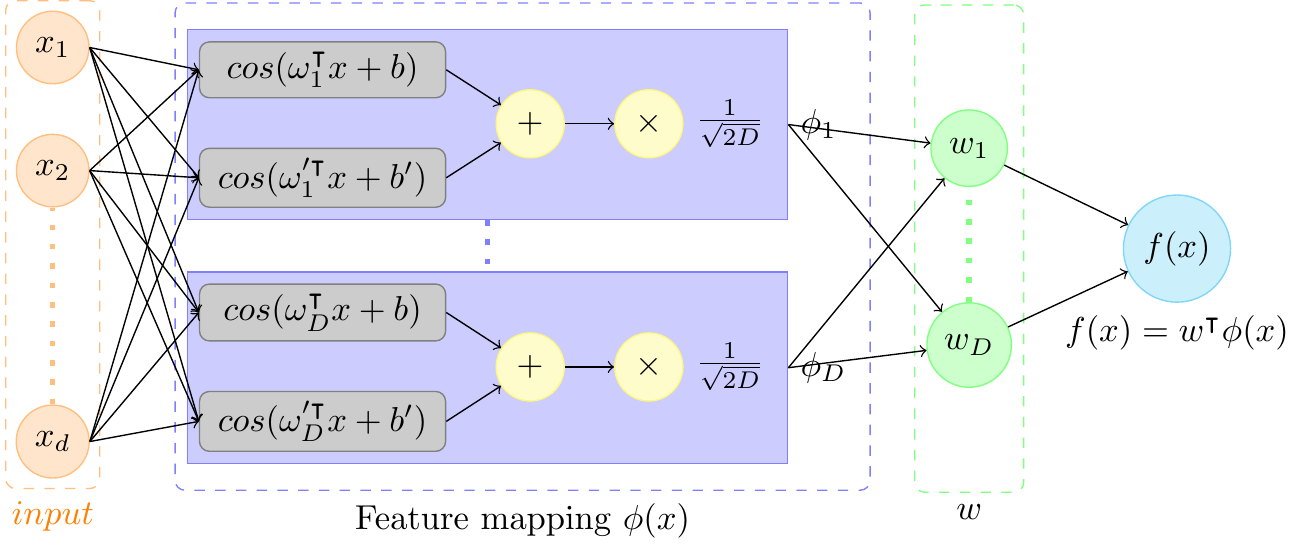}
    \caption{Architecture of ASKL critic. The feature mapping $\phi$ is parameterized by the middle layer. Its weights are the frequency samples $\left\{\omega_i,\omega_i^\prime\right\}_{i=1}^{D}$ sampled from spectral distribution $S\left(\omega,\omega^\prime\right)$. The output layer parameterizes the RKHS representation $w$ of a function $f$ such that $f\left(x\right) = w^\intercal\phi\left(x\right)$ }. 
    \label{fig:critic_figure}
\end{figure}
In this subsection we discuss Reproducing Hilbert Kernel Spaces (RKHS) and Automated Spectral Kernel Learning (ASKL). \textit{Hilbert space} $\mathcal{H}$, is an vector space of real valued functions $f:\mathcal{X}\rightarrow\mathbb{R}$ with defined inner product $\langle f,g  \rangle_{\mathcal{H}}$ between two functions $f$ and $g$. Function norm in the hilbert space is defined as $\left\lVert f \right\rVert_\mathcal{H} = \sqrt{\langle f,f \rangle_\mathcal{H}}$.
\textit{Reproducing kernel} of a hilbert space is a positive semi-definite function, $K:\mathcal{X}\times\mathcal{X}\rightarrow\mathbb{R}$ which satisfies the conditions: (1) $K\left(\cdot,x\right) \in \mathcal{H}\; \forall x \in \mathcal{X}$, and (2) $\langle f,K\left(\cdot,x\right) \rangle_\mathcal{H} = f\left(x\right)\; \forall f \in \mathcal{H}\; \&\; \forall x \in \mathcal{X}$. The latter of the two condition is known as the reproducing property of the kernel $K$ \cite{berlinet2011reproducing}. A Hilbert space which posses a reproducing kernel is called a \textit{Reproducing Kernel Hilbert Space}.

There exist many feature mappings, $\varphi:\mathcal{X}\rightarrow\mathcal{F}$, where $\mathcal{F}$ is a Hilbert space, such that $K(x,y) = \langle \varphi(x), \varphi(y) \rangle_{\mathcal{F}}$  and $f(x) = \langle w,\varphi(x) \rangle_\mathcal{F}$, $w \in \mathcal{F}$, and $f \in \mathcal{H}$. A special case of such feature mappings known as implicit feature mapping is $\phi(x) = K\left(\cdot,x\right)$ and $K\left(x,y\right) = \langle \phi\left(x\right), \phi\left(y\right) \rangle_\mathcal{H}$.

Yaglom's theorem \cite{yaglom1987correlation} as stated below shows that there exists a duality between a positive semidefinite kernel function and a non-negative Lebesgue-Stieltjes measure in spectral domain.     
\begin{theorem}
\label{th:yagloms}(Yaglom's theorem) A kernel $K(x,y)$ is positive semi-definite \textit{iff} it can be expressed as
\begin{equation*}
\label{eq:yagloms}
K(x,y) = \int_{\mathcal{R}^d\times\mathcal{R}^d} e^{i(\omega^\intercal x - \omega^{\prime\intercal} y)} dS(\omega,\omega^\prime)
\end{equation*}
where, $S(\omega,\omega^\prime)$ is Lebesgue-Stieltjes measure associated to some positive semi-definite function $s(\omega,\omega^\prime)$ with bounded variations.
\end{theorem}

With appropriate scaling the Lebesgue-Stieltjes measure $S\left(\omega,\omega^\prime\right)$ can be treated as a probability distribution in spectral domain where $\omega$ and $\omega^\prime$ are spectral variables. From here on, this probability distributions is referred to as spectral distribution. An implication of theorem \ref{th:yagloms} is
that it is possible to learn an RKHS associated with a kernel by learning a spectral distribution. 

Automated Spectral Kernel Learning (ASKL) \cite{li2019automated} is a kernel learning method that used samples from the spectral distribution $S\left(\omega,\omega^\prime\right)$ to construct a feature mapping $\phi\left(x\right)$ defined as,
\begin{equation}
    \phi(x) = \frac{1}{\sqrt{2D}} [cos(\Omega^\intercal x + b) + cos(\Omega^{\prime\intercal}x + b^\prime)]
\end{equation}
Where, $\Omega = \left[\omega_1,\dots,\omega_D\right]$ and $\Omega^\prime = \left[\omega_1^\prime,\dots,\omega_D^\prime\right]$ are $d\times D$ matrices of frequency samples $\{\omega_i,\omega_i^\prime\}_{i=1}^D \stackrel{iid}{\sim}S(\omega,\omega^\prime)$ and $b$ and $b^\prime$  are vectors of $D$ uniform samples $\{b_i\}_{i=1}^D,\{b_i^\prime\}_{i=1}^D \stackrel{iid}{\sim} \mathcal{U}[0,2\pi]$. The kernel associated with the spectral distribution can be approximated using the feature mapping $\phi\left(x\right)$ defined above as $K\left(x,y\right) = \phi\left(x\right)^\intercal \phi\left(y\right)$. This feature mapping $\phi\left(x\right)$ produces a $D$-dimensional embedding in an RKHS for any input $x$. Any function in this RKHS is represented by a $D$-dimensional vector $w$, such that $f\left(x\right) = w^\intercal \phi\left(x\right)$.

ASKL represented the RKHS generated by the above feature mapping as a two layer neural network with cosine activations shown in Fig. \ref{fig:critic_figure}. The hidden layer of this neural network represents the feature mapping $\phi\left(x\right)$, its trainable parameters are the frequency samples $\left\{\omega_i,\omega_i^\prime\right\}$ from spectral distribution $S\left(\omega,\omega^\prime\right)$. The parameters $w$ of the final output layer represent functions in the RKHS. The output of the final layer is the inner product $f\left(x\right) = \langle w,\phi\left(x\right) \rangle_\mathcal{H}$. A RKHS can be learned by optimizing this neural network using a stochastic gradient descent method. During the optimization, a spectral distributions is learned implicitly through learning the parameters of the hidden layer $\left\{\omega_i,\omega_i^\prime\right\}$. In this work, the critic's hypothesis space is restricted to an RKHS using the  neural network architecture Fig. \ref{fig:critic_figure} and ASKL. For more information on ASKL refer to \cite{li2019automated}. Any further reference to ASKL critic refers to the neural network architecture shown in Fig. \ref{fig:critic_figure}.

\section{Theory \& Our Approach}
\label{our_approach}
Our goal is to estimate the mutual information, $I(X;Y)$, between two RVs $X$ and $Y$, from $n$ i.i.d samples, $\left\{x_i,y_i\right\}_{i=0}^n$ from joint distribution $\mathbb{P}_{XY}$ and $m$ i.i.d samples, $\left\{x_i^\prime,y_i^\prime\right\}_{i=0}^m$ from the product of marginal distributions $\mathbb{P}_X\otimes\mathbb{P}_Y$. As, the true underlying probability distributions are unknown, we use empirical approximations of the variational lower bounds of MI defined as:

\begin{align}
    \hat{I}_{TUBA}^{n,m}\left(f_\theta,S\right) &= \mathbb{E}_{\mathbb{P}_{XY}^n}\left[f_\theta\left(x,y\right)\right] - \nonumber \\ 
    &\quad\frac{\mathbb{E}_{\mathbb{P}_X^m\otimes\mathbb{P}_Y^m}\left[e^{f_\theta\left(x,y\right)}\right]}{a} -log\left(a\right) + 1 \label{eq:empirical_tuba}\\
    \hat{I}_{DV}^{n,m}\left(f_\theta,S\right) &= \mathbb{E}_{\mathbb{P}_{XY}^n}\left[f_\theta\left(x,y\right)\right] - \nonumber \\ 
    &\quad log\left(\mathbb{E}_{\mathbb{P}_X^m\otimes\mathbb{P}_Y^m}\left[e^{f_\theta\left(x,y\right)}\right]\right)
\end{align}

Where, $S$ is the set of n,m i.i.d samples$\left\{x_i,y_i\right\}_{i=1}^n$. $\left\{x_i^\prime,y_i^\prime\right\}_{i=1}^m$, $\mathbb{P}_{XY}^n$ and $\mathbb{P}_X^m\otimes\mathbb{P}_Y^m$ are empirical distributions corresponding to samples $\left\{x_i,y_i\right\}_{i=1}^n$ and $\left\{x_i^\prime,y_i^\prime\right\}_{i=1}^m$, respectively, $\mathbb{E}_{\mathbb{P}_{XY}^n}\left[f\left(x,y\right)\right] = \frac{1}{n}\sum_{i=1}^n f\left(x_i,y_i\right)$ and $\mathbb{E}_{\mathbb{P}_{X}^m\otimes\mathbb{P}_Y^m}\left[f\left(x,y\right)\right] = \frac{1}{m}\sum_{i=1}^m f\left(x_i^\prime,y_i^\prime\right)$.

\begin{figure*}
    \centering
    \subfigure[Comparison on 20 dimensional correlated Gaussian dataset]{
        \includegraphics[width=0.9\textwidth]{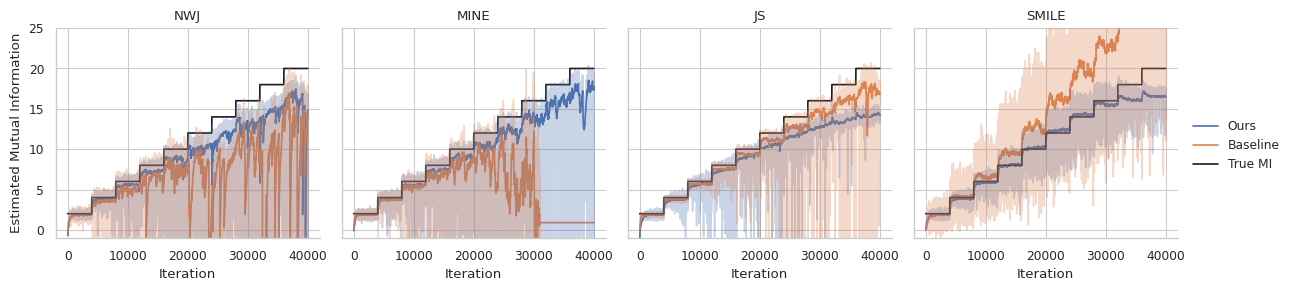}
        \label{fig:mi_plot_non_cube}
    }
    \subfigure[Comparision on cubed 20 dimensional correlated Gaussian dataset]{
        \includegraphics[width=0.9\textwidth]{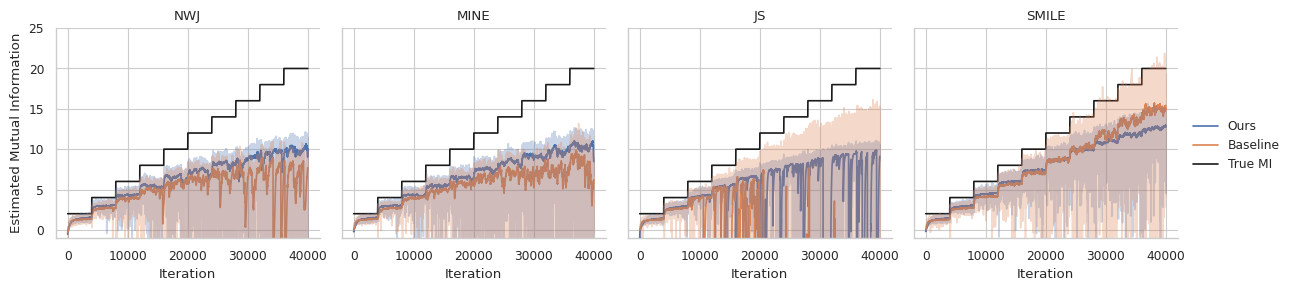}
        \label{fig:mi_plot_cube}
    }
    \caption{Qualitative comparison between ASKL and baseline critic on four diferent variational lower bounds of MI, $I_{NWJ}$, $I_{MINE}$, $I_{JS}$, and $I_{SMILE}$. MI estimates on Gaussian correlated and cubed Gaussian correlated datasets are plotted in \subref{fig:mi_plot_non_cube} and \subref{fig:mi_plot_cube}, respectively. MI estimate by the proposed ASKL critic are in blue and the estimates of baseline critic are depicted in orange. The solid plotted lines are exponentially weighted moving average of these estimates. ASKL critic etimates are more stable in comparison to baseline estimates on all lower bounds of MI and both datasets. A specific case of estimation instability can be noticed in $I_{MINE}$ (first row second plot) based estiamtion of MI using baseline critic architecture when the true MI is higher than 16, whereas, ASKL critic computes stable MI estimates even at higher values.}
    \label{fig:mi_plots}
\end{figure*}

\subsection{Theoretical Guarantees}
In this subsection the generalization behaviour of the empirical estimates, $\hat{I}_{TUBA}^{n,m}$ and $\hat{I}_{DV}^{n,m}$ are discussed. We derive generalization error bound for the empirical estimates using data-driven Rademacher complexity of general critic's hypothesis space. We also bound the empirical Rademacher complexity of the ASKL critic's hypothesis space.

Generalization error quantifies the out of sample behaviour of an estimator. Formally, generalization error is defined as the maximum possible deviation of the empirical estimates from true values. If empirical estimate $\hat{I}$ is an unbiased estimate, then variance of this empirical estimate is upper bounded by the expectation of squared generalization error. Hence, generalization error is an indicator of the variance of the estimate. The following theorem bounds the generalization error of $\hat{I}_{TUBA}^{n,m}$ and $\hat{I}_{DV}^{n,m}$.

\begin{theorem}[Generalization Error Bounds]
\label{theorem:generalization_bounds}
Assume, that the hypothesis space $\mathcal{F}$ of the critic is uniformly bounded by $M$, that is  $\left|f(x,y)\right| \leq M\; \forall f \in \mathcal{F}\;\&\; \forall \left(x,y\right) \in \mathcal{X}\times\mathcal{Y}$, $M < \infty$. For a fixed $\delta >0$ generalization errors of 
$\hat{I}_{TUBA}^{n,m}$ and $\hat{I}_{DV}^{n,m}$ can be bounded with probability of at least $1-\delta$, given by

\begin{multline}
    \label{eq:tuba_generalization}
    \sup\limits_{f \in \mathcal{F}} \left(I_{TUBA}(f) - \hat{I}_{TUBA}^{n,m}(f)\right) \leq 4 \hat{\mathcal{R}}_n\left(\mathcal{F}\right) + \frac{8}{a} e^M \hat{\mathcal{R}}_m\left(\mathcal{F}\right)  \\ \quad\quad+\frac{4M}{n}log\left(\frac{4}{\delta}\right) +
    \frac{8Me^M}{am}log\left(\frac{4}{\delta}\right) \\
    +\sqrt{\frac{\left(\frac{4M^2}{n} + \frac{\left(e^M-e^{-M}\right)^2}{a^2m}\right) log\left(\frac{2}{\delta}\right)}{2}}
\end{multline}

\begin{multline}
    \label{eq:dv_generalization}
    \sup\limits_{f \in \mathcal{F}}\left(I_{DV}(f) - \hat{I}_{DV}^{n,m}(f)\right) \leq 4\hat{\mathcal{R}}_n\left(\mathcal{F}\right) + 8e^{2M}\hat{\mathcal{R}}_m\left(\mathcal{F}\right)\\ + \frac{4M}{n} log\left(\frac{4}{\delta}\right) + 
    \frac{8Me^{2M}}{m} log\left(\frac{4}{\delta}\right)\\ + \sqrt{\frac{\left(\frac{4M^2}{n} + \frac{\left(e^{2M}-1\right)^2}{m}\right) log\left(\frac{2}{\delta}\right)}{2}}
\end{multline}

Where, sample set $S$ for $\hat{I}_{TUBA}^{n,m}$ and $\hat{I}_{DV}^{n,m}$ is assumed to be known, and $\hat{\mathcal{R}}_n\left(\mathcal{F}\right)$ and $\hat{\mathcal{R}}_m\left(\mathcal{F}\right)$ are empirical Rademacher averages of the hypothesis space $\mathcal{F}$ for different sample sizes.
\end{theorem}

To formulate the generalization error bounds given in the above theorem, we used McDairmid's inequality to bound generalization error by expected generalization error over sample set $S$.  
Then we use lemma A5 given in \cite{bartlett2005local} to bound the expected error by Rademacher complexity. Further, Rademecher concentration inequality, lemma A4 also given in \cite{bartlett2005local} is used to arrive at the final theoretical guarantees. Refer to Appendix B for detailed proof. Error bounds for $I_{NWJ}$ and $I_{MINE}$ are derived by substituting the parameter $a$ in bound \ref{eq:tuba_generalization} with $e$, and with exponential moving average of $e^f_\theta\left(x,y\right)$ across mini-batches, respectively. $I_{JS}$ uses $I_{NWJ}$ lower bound to estimate MI, hence generalization error of $I_{JS}$ is bounded by generalization error bound of $I_{NWJ}$. Similarly, $I_{SMILE}$ uses $I_{DV}$ lower bound to estimate MI and its generalization error is bounded by error bound of $I_{DV}$ 

The generalization error bounds depend on the empirical Rademacher complexities and $e^M$. Our finding on the dependence of the generalization error on $e^M$ is confirmed by similar observation made in \cite{mcallester2018formal} on the sample complexity of MINE estimator. From the error bounds, it can be inferred that high empirical Rademacher complexity of the critic's hypothesis space leads to high generalization error, hence high variance estimates. Therefore, variance of these estimates can be effectively reduced by choosing a hypothesis space for critic with low Rademacher complexity. However, it is also necessary to keep the hypothesis space rich enough to induce low bias. Though these bounds apply to all hypothesis spaces including the space of functions that are learned by a fully connected neural network, emperical estimation of Rademacher complexity for a fully connected neural network is an open area of research. We restrict the critic neural networks hypothesis space to RKHS by using ASKL to gain insights into variational lower bound estimates of MI. The empirical Rademacher complexity of the ASKL critic's hypothesis space can be upper bounded as shown by the following theorem, 

\begin{figure*}
    \centering
    \subfigure[Bias, variance and RMSE of ASKL critic estimates for different batch sizes.]{
        \label{fig:askl_batch_size_bias_variance}
        \includegraphics[width=0.45\textwidth]{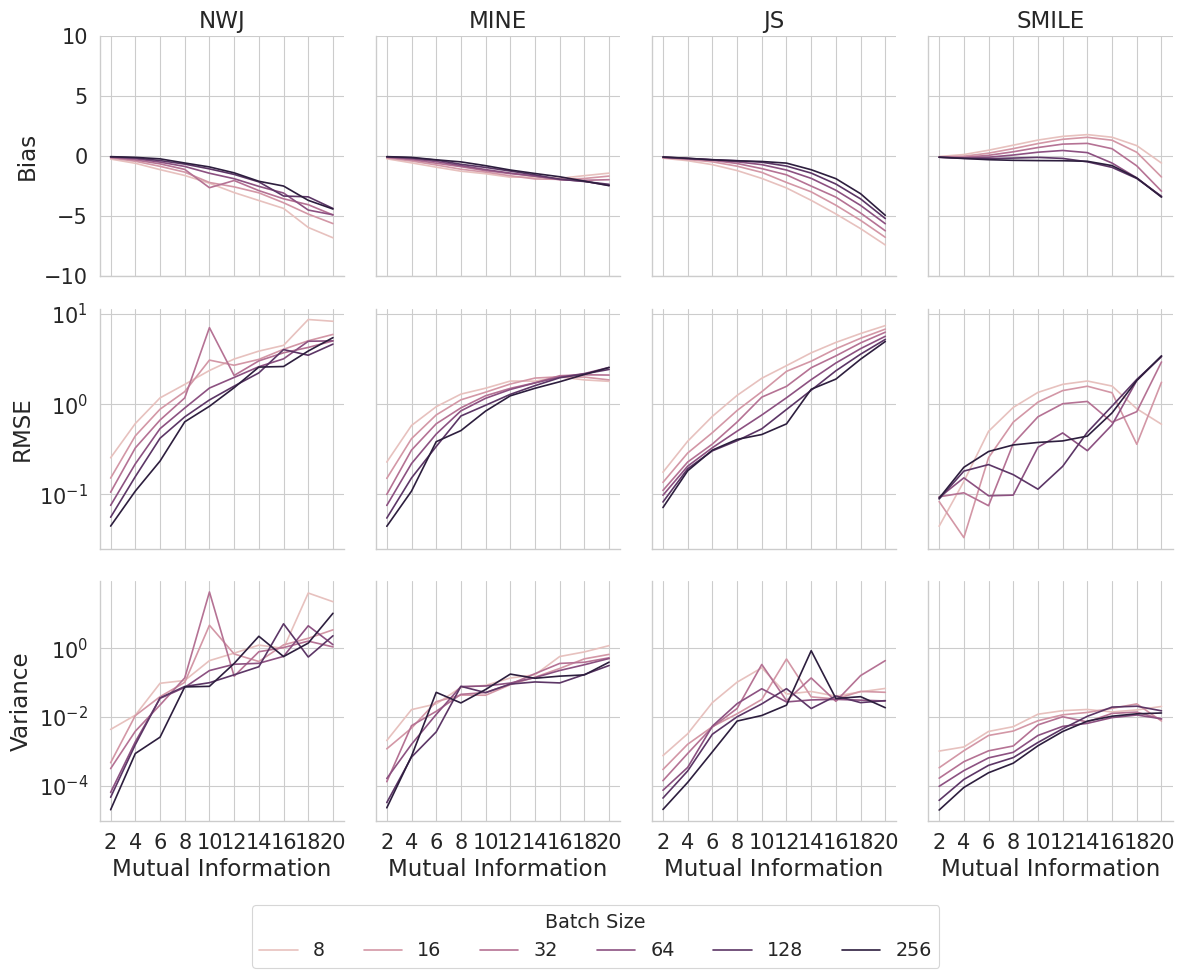}
    }
    \subfigure[Bias, variance and RMSE of baseline critic estimates for different batch sizes.]{
        \label{fig:baseline_batch_size_bias_variance}
        \includegraphics[width=0.45\textwidth]{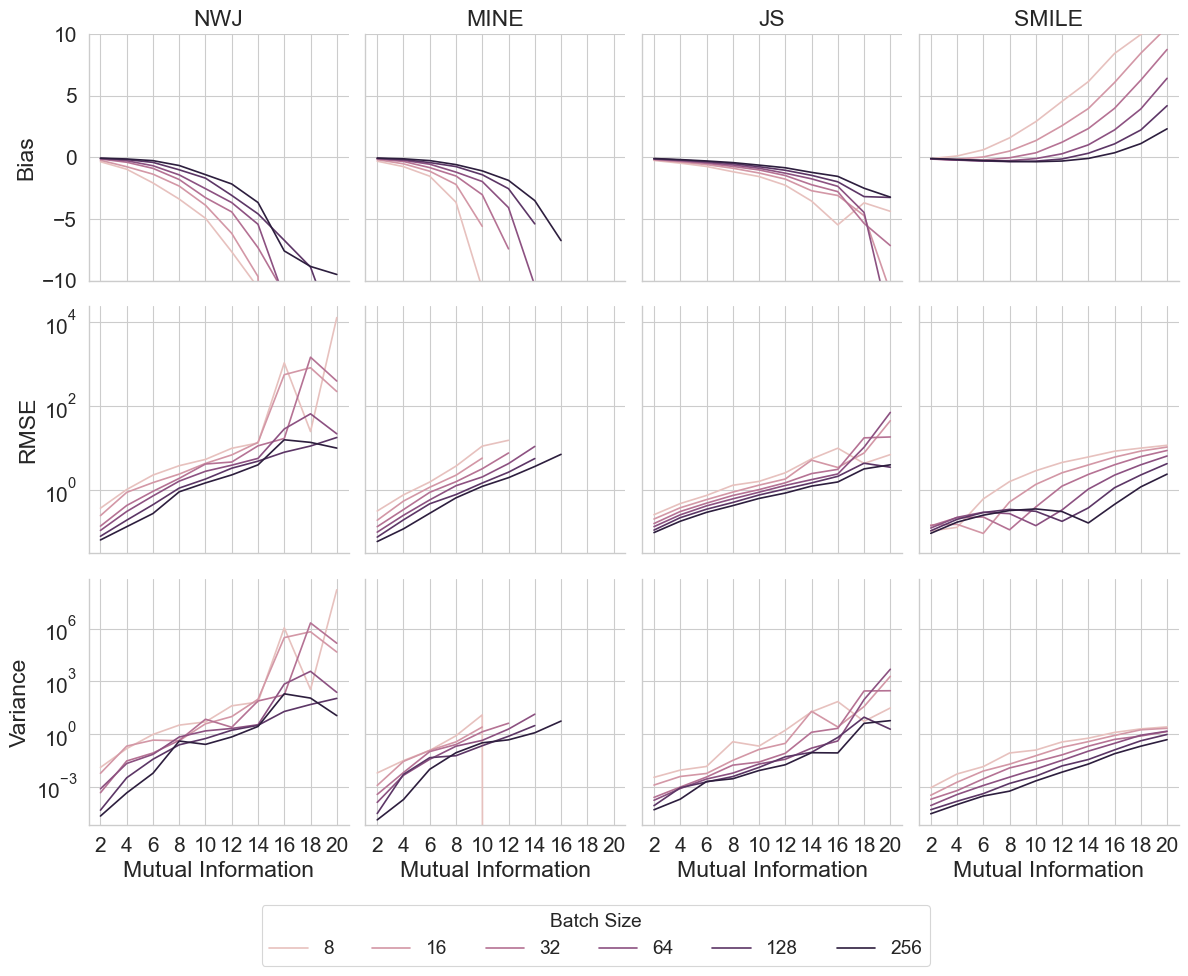}
    } \quad
    \caption{Bias, variance, and RMSE values of ASKL critic and baseline critic estimates averaged over 50 experimental trials are shown in figures \protect\subref{fig:askl_batch_size_bias_variance} and \protect\subref{fig:baseline_batch_size_bias_variance}, respectively. In each figure first, second and third rows contain bias, RMSE and variance plots. Each column corresponds to different lower bound, and in each plot different plotted lines correspond to different batch sizes. ASKL critic etimates are less biased and exhibit lower variance compared to baseline critic estimates on all variational lower bounds.}
    \label{fig:batch_size_bias_variance_rmse}
\end{figure*}
\begin{theorem}
    \label{theorem:askl_rademacher}
    The empirical Rademacher average of the RKHS $\mathcal{F}$ to which ASKL critic belongs can be bounded as following
    \begin{align*}
        \hat{\mathcal{R}}_n(\mathcal{F}) &\leq \frac{B}{n} \sqrt{\sum_{i=1}^n\lVert \phi\left(x_i\right)\rVert_2^2} \leq \frac{B}{\sqrt{n}}
    \end{align*}
    Where $B = \sup\limits_{f\in\mathcal{F}}\lVert w \rVert_2$.
\end{theorem}

We used the Cauchy-Schwarz inequality to bound the complexity of the ASKL critic, for detailed proof refer to Appendix A. Note that, the second inequality in the above theorem is true only in the case of ASKL critic. Using the above theorem we can decrease the complexity by decreasing the largest possible norm of RKHS representation of functions $w$ or decreasing the frobenius norm of the feature mapping matrix. In the next subsection, we present an optimization procedure to decrease the empirical Rademacher complexity by penalizing $\lVert w \rVert_2$ and $\lVert \phi\left(X\right) \rVert_F$ to control the bias-variance tradeoff. Using second inequality, and penalizing $\lVert w \rVert_2$ it is possible to carve out the regularisation used by Nguyen \etal \cite{nguyen2010estimating} to control hypothesis space complexity.

\subsection{Training Methodology}
We train an ASKL critic neural network shown in Fig. \ref{fig:critic_figure} to simultaneously maximize empirical estimate of MI and minimize regularization terms defined below. The overall training objective is:
\begin{equation}
    \underset{\theta}{\mathrm{argmin}} -\hat{I}\left(f_\theta,S\right) + \lambda_1 \lVert w \rVert_2 + \lambda_2 \lVert \phi\left(S;\theta\right)\rVert_F
\end{equation}

Where, $\hat{I}$ can be an empirical estimate of any variational lower bound of MI, $\hat{I}_{NWJ}^{n,m}$, $\hat{I}_{MINE}^{n,m}$, $\hat{I}_{JS}^{n,m}$ or $\hat{I}_{SMILE}^{n,m}$. And $\theta$ is the set of trainable parameters $w$, $\Omega$, and $\Omega^\prime$. GAN discriminator objective is maximized in cases where $\hat{I}$ is $\hat{I}_{JS}^{n,m}$ or $\hat{I}_{SMILE}^{n,m}$. In this work, regularization terms $\lVert w \rVert_2$ and $\lVert \phi\left(S; \theta\right) \rVert_F$ appear in upper bound of empirical Rademacher complexity of ASKL critic's hypothesis space. 
Bias-variance tradeoff is controlled by tuning hyperparameters, $\lambda_1$ and $\lambda_2$. We use mini-batch stochastic gradient decent to train the estimator.

\section{Experiments}
\label{experiments}
\begin{table}
    \centering
    \caption{Regularization weights}
    \begin{tabular}{|c|c|c|}
        \hline
        Lower Bound & $\lambda_1$ & $\lambda_2$\\
        \hline
        \hline
        NWJ\cite{nowozin2016f} & 0.001 & 0.001 \\
        MINE\cite{belghazi2018mutual} & 0.001 & 0.001\\
        JS\cite{poole2018variational} & 1e-5 & 1e-5\\
        SMILE\cite{song2019understanding} & 1e-4 & 0.001\\
        \hline
    \end{tabular}
    \label{tab:lambda_table}
\end{table}
We empirically validate our claims on two different toy datasets which have been widely used by other MI estimation methods \cite{belghazi2018mutual,poole2018variational,song2019understanding}, (1) correlated Gaussian dataset, where samples of two RVs $\left(X,Y\right)$ are drawn from a 20 dimensional Gaussian distribution with correlation $\rho$ between each dimension of $X$ and $Y$. The correlation $\rho$ is increased such that $I\left(X;Y\right)$ increases in steps of 2 every 4000 training steps, and (2) cubed Gaussian dataset, same as in (1) but we apply a cubic non-linearity to $Y$ to get samples ($x$, $y^3$). As, mutual information remains unchanged by application of deterministic functions on random variables, $I\left(X;Y^3\right) = I\left(X;Y\right)$.
Further, is it important to note that previous methods increased the correlation $\rho$ till the true MI is increased to 10. In our experimental analysis, we increased the correlation $\rho$ till the true MI is 20 to demonstrate that ASKL critic produces low variance estimates even at high values of MI. 

For comparative analysis we train ASKL critic and a baseline critic on four different lower bounds, namely $I_{NWJ}$, $I_{MINE}$, $I_{JS}$, and $I_{SMILE}$. The baseline critic is a fully connected neural network with ReLU activations. This baseline has been used by previous estimation methods that consider the universal approximation property of neural networks \cite{belghazi2018mutual,poole2018variational,song2019understanding}. ASKL critic with regularised space complexity computes low variance stable variational lower bound estimates of MI in comparison to baseline critic. 

Code for this paper are available at \url{https://cvit.iiit.ac.in/projects/mutualInfo/}.

\begin{figure*}
    \centering
    \subfigure[Bias of ASKL critic based estimators for different configurations of regularization weights]{
        \label{fig:askl_regularization_bias}
        \includegraphics[width=0.45\textwidth]{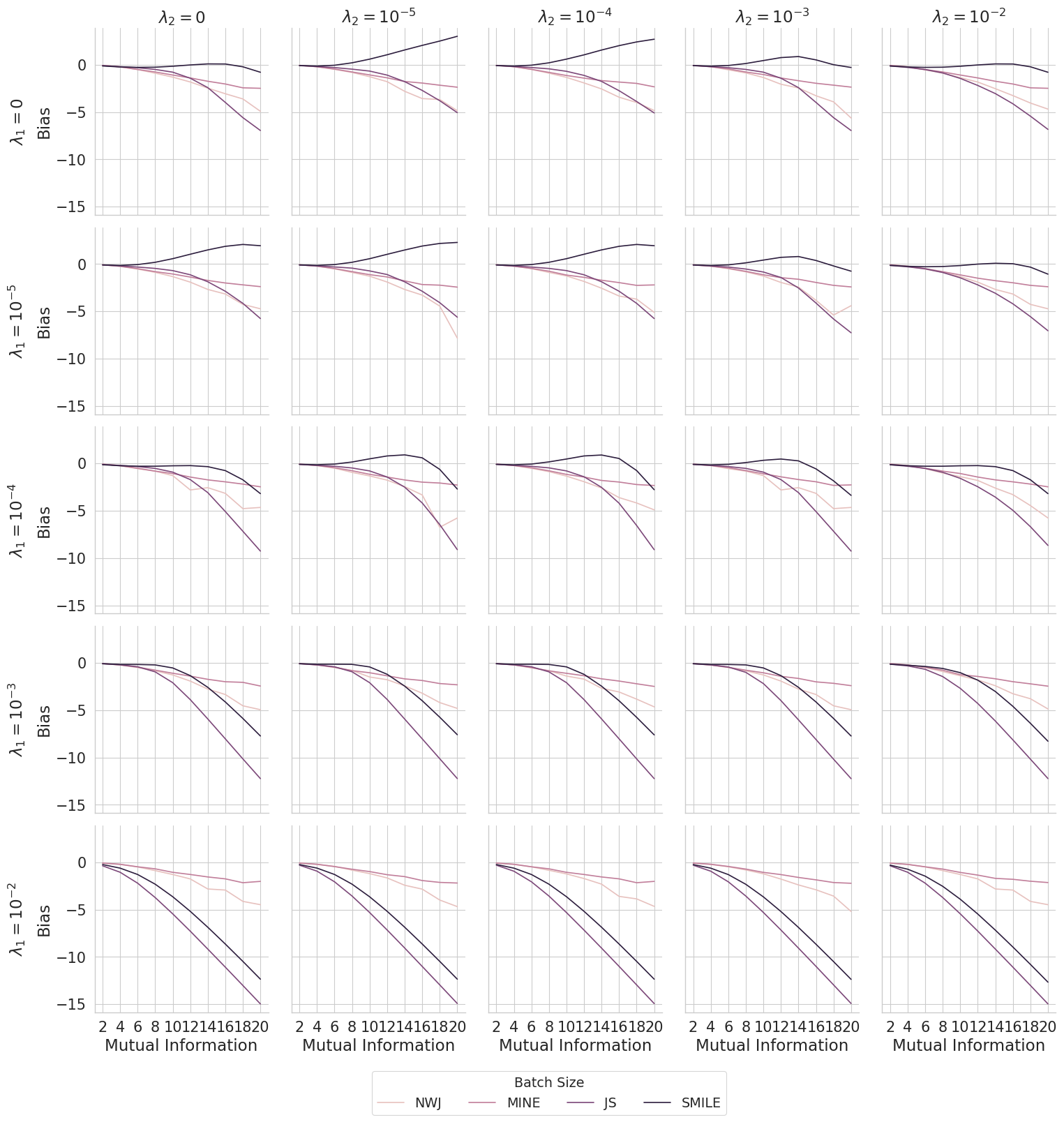}
    }
    \subfigure[Variance of ASKL critic based estimators for different configurations of regularization weights]{
        \label{fig:askl_regularization_variance}
        \includegraphics[width=0.45\textwidth]{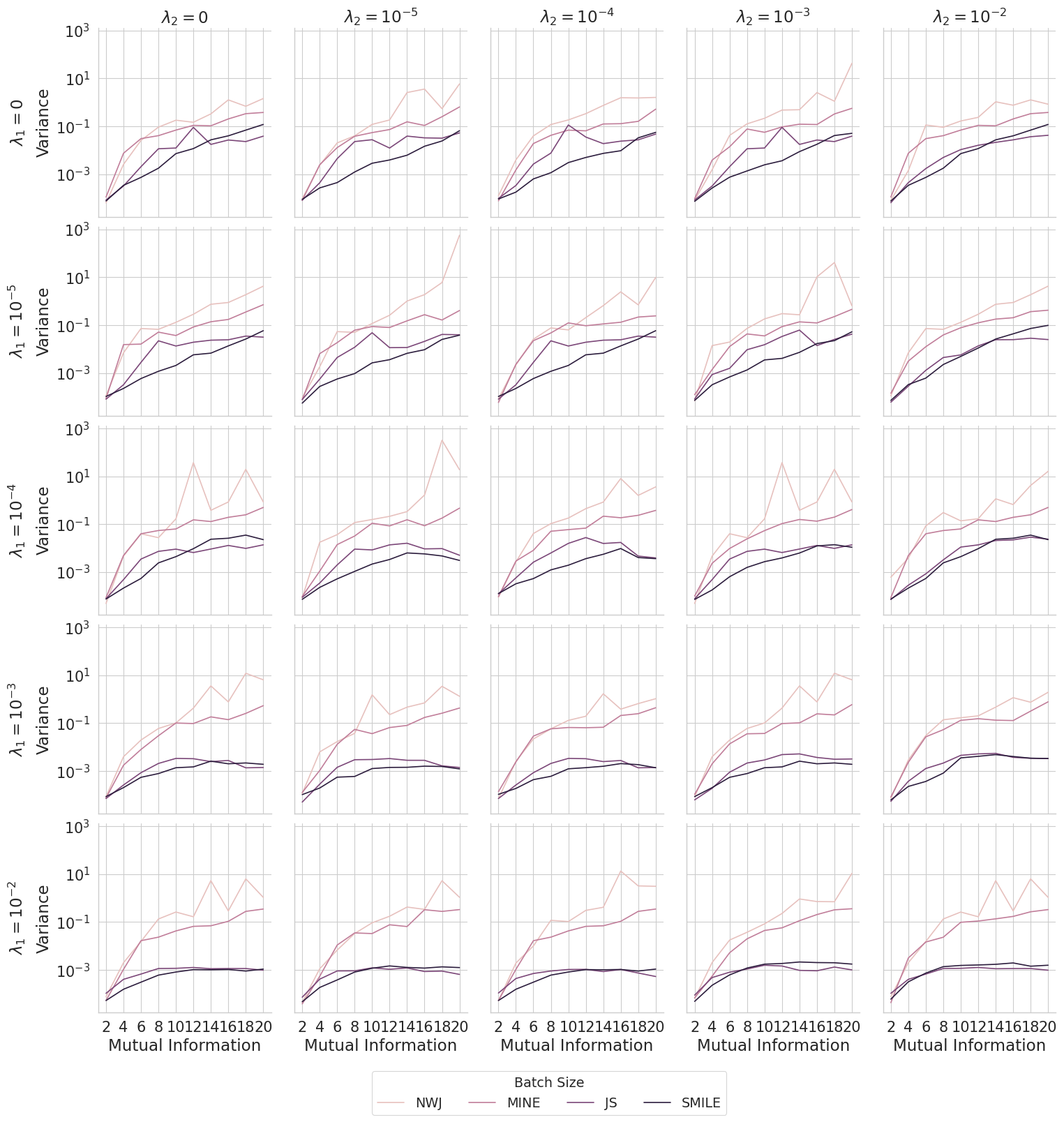}
    }
    \caption{Bias-variance tradeoff for different values of $\lambda_1$ and $\lambda_2$ for estimation using ASKL critic. Figures \protect\subref{fig:askl_regularization_bias} and \protect\subref{fig:askl_regularization_variance} show bias and variance plots, respectively. In both figures each row corresponds to a single $\lambda_1$ value and each column corresponds to a single $\lambda_2$ value. These figures quantitatively demonstrate the efficacy of the proposed regularisation terms ($\lambda_1$ and $\lambda_2$ for $\lVert w \rVert_2$ and $\lVert \phi\left(X\right) \rVert_F$, respectively) in controlling bias-variance tradeoff of ASKL critic's space complexity.}
    \label{fig:regularization_bias_variance}
\end{figure*}
\subsection{Training Details}
For ASKL critic, $D$ is set to 512, that is 512 spectral samples are used for estimation. The multiplicity factors for each of the regularization terms used for different estimators are given in Table \ref{tab:lambda_table}. For our baseline critic, we used a 3 layer neural network with 256 units in each hidden layer. Unless mentioned otherwise, batch size is set to 64. We use Adam optimizer \cite{kingma2014adam} with $\beta_1 = 0.9$ and $\beta_2=0.999$. Learning rates are set to $10^{-3}$ and $5\times10^{-4}$ for ASKL and baseline critics, respectively.

We test the validity of our claim that constraining the critic to RKHS should lead to better bias-variance tradeoff in three different experimental setups, (1) qualitatively compare the variance of MI estimates between ASKL critic and baseline critic on four different variational lower bounds of MI. These experiments are performed on both toy datasets described above, batch size is fixed at 64 sample, (2) quantitatively compare the average bias, variance, and the root mean square error (RMSE) between the true and empirical estimates of MI over 50 experimental trials. These quantitative comparisons are made over a range of batch sizes to depict the robustness of our estimates with varying batch sizes, (3) quantitatively demonstrate the efficacy of the proposed regularisation terms in controlling bias-variance tradeoff of ASKL critic's space complexity by varying the regularisation hyperparameters $\lambda_1$ and $\lambda_2$ for $\lVert w \rVert_2$ and $\lVert \phi\left(X\right) \rVert_F$, respectively. In experiment (3), bias-variance values are estimated over 50 experiments. Both experients (1) and (2) are run on correlated Gaussian dataset. We further elaborate on each of these experimental results in the next subsection.   

\subsection{Results}

Qualitative comparison between ASKL critic and baseline critic on four different variational lower bounds of MI has been shown in Fig. \ref{fig:mi_plots}. Fig. \ref{fig:mi_plot_non_cube} and Fig. \ref{fig:mi_plot_cube} demonstrate the comparative results on the 20 dimensional correlated Gaussian dataset and the cubed correlated Gaussian dataset, respectively. In can be seen that maximisation using ASKL critic tends to produce stable estimates in comparison to their baseline counterpart. A particular instance of numerical instability in baseline critic estimates can be observed in the plot corresponding to $I_{MINE}$ when the true MI is higher than 16. Estimates by ASKL critic does not suffer from such instability and it is to be noted that the ASKL critic also produces comparatively low variance MI. 

We compute bias, variance, and root mean square error of the estimated MI values to quantitatively evaluate the proposed ASKL critic's performance against the baseline. The bias, variance, and RMSE values have been averaged over 50 experimental trials. Fig. \ref{fig:askl_batch_size_bias_variance} and Fig. \ref{fig:baseline_batch_size_bias_variance} show the computed values for the ASKL critic and the baseline, respectively. These plots conclusively demonstrate that the ASKL critic estimates have lower bias and variance characteristics in comparison to the baseline critic. Lower variance characteristics of the ASKL critic can be explained by observing that the empirical Rademacher complexity of ASKL critic's hypothesis space is bounded, theorem \ref{theorem:askl_rademacher}. Hence, generalization error is guaranteed to be upper bounded. Lower bias in estimates can be attributed to better control over bias-variance tradeoff. 

Experimental results shown in Fig. \ref{fig:batch_size_bias_variance_rmse}, demonstrates the effect of change in batch size on the variance of ASKL and baseline critic estimates. It can be observed that with an increase in batch size the variance of both ASKL and baseline estimates decreases. This is due to the fact that the empirical Rademacher complexity is inversely proportional to the sample size (refer Appendix A for definition). Hence, an increase in batch size leads to a decrease in empirical Rademacher complexity and, corresponding decrease in variance of the MI estimates. Another key observation on the variance of MI estimates which holds for both critics is that with an increase in true MI the variance of the empirical estimates increases. This observations can be explained by noticing the effect of increase in the value of true MI on the log likelihood density ratio between the joint and product of marginal distributions, $log\left(d\mathbb{P}_{XY}/d\mathbb{P}_X\otimes\mathbb{P}_Y\right)$. The absolute value of the log density ratio evaluated at any given sample increases with increase in MI. The optimal critics for variational lower bound estimates of MI depend on the log density ratio. Hence, to match the increase in log density ratio the constant $M$ which uniformly bounds the critic's hypothesis space also increases. As described in theorem \ref{theorem:generalization_bounds}, the generalization error bounds depend on both empirical Rademacher complexity and $e^M$, hence, an increase in $M$ leads to an increase in variance of MI estimates.

Bias-variance tradeoff for different values of $\lambda_1$ and $\lambda_2$ in ASKL critic, figure \ref{fig:regularization_bias_variance}. Figures \ref{fig:askl_regularization_bias} and \ref{fig:askl_regularization_variance} are the bias and variance plots, respectively. The left top most plots in both figures, \ref{fig:askl_regularization_bias} and \ref{fig:askl_regularization_variance} correspond to $\lambda_1$ and $\lambda_2$ set to 0, respectively. It can be seen in these plots that even without any explicit regularisation estimates using ASKL critic have lower bias and lower variance in comparison to the baseline critic. This verifies our claim that constraining the complexity of the hypothesis space leads to significant improvement in reliability of these estimates. It is evident from these plots that regularization weights are also effective in controlling the bias, as $\lambda_1$ and $\lambda_2$ increase the estimates get biased in negative direction. This demonstrates the efficacy of the proposed regularization terms in inducing effective bias-variance tradeoff.

\section{Conclusion}
\label{conclusion}
In the proposed work, we successfully demonstrate the effect of controlling the complexity of critic's hypothesis space on the variance of sample based empirical estimates of mutual information. We negate the high variance characteristics of variational lower bound based estimates of MI by constructing the critic's hypothesis space in a Reproducing Kernel Hilbert Space, which corresponds to a critic learned using Automated Spectral Kernel Learning architecture. By analysing the generalisation bounds using Radmacher complexity of the constrained critic space, we demonstrate effective regularisation of bias-variance tradeoff on four different variational lower bounds of Mutual information. In larger scheme of Explainable-AI, this work theoretically motivates the implications of understanding the effect of regulating the complexity of deep neural network based critic hypothesis spaces on the bias-variance tradeoff of variational lower bound estimators of mutual information.

{\bibliographystyle{IEEEtran}
\bibliography{ms}
}
\end{document}

\maketitle

\appendix
\section{Rademacher Complexity}
In problems pertinent to machine learning obtaining practical generalization error bound is crucial for proper model selection. Generalization error bounds are typically contained by a measure of the complexity of the learning model's hypothesis space, for example, the covering number of the hypothesis function space. The data-driven Rademacher's complexity used in this work is described as follows:

Let $(\mathcal{X},\mathbb{P})$ be a probability space and $\mathcal{F}$ be the class of measurable functions from $\mathcal{X}$ to $\mathbb{R}$. Consider $X_1,X_2,\dots,X_n$ to be $n$ i.i.d data samples from $\mathbb{P}$, with the corresponding empirical distribution denoted by $\mathbb{P}_n$. Now, let $\sigma_1,\sigma_2,\dots,\sigma_n$ be $n$ independent discrete random variables for which $Pr(\sigma=1)=Pr(\sigma=-1)=\frac{1}{2}$ known as the Rademacher random variables. Then, for any $f\in\mathcal{F}$ we define
\begin{equation}
\begin{gathered}
R_n(f) = \frac{1}{n}\sum_{i=1}^{n}\sigma_i f(X_i),\; R_n{\mathcal{F}} = \sup\limits_{f\in\mathcal{F}}R_n{f}\\ \hat{\mathcal{R}}_{n}(\mathcal{F}) = \mathbb{E}_\sigma\left[R_n(\mathcal{F})\right],\;\mathcal{R}_n(\mathcal{F}) = \mathbb{E}\left[R_n(\mathcal{F})\right] 
\end{gathered}
\end{equation}

Where, $\mathbb{E}_\sigma$ denotes expectation with respect to the Rademacher random variables, $\left\{\sigma_i\right\}_{i=1}^n$. And $\mathbb{E}$ is the expectation with respect to Rademacher random variables and data samples, $\left\{X_i\right\}_{i=1}^n$. 
$\mathcal{R}_n(\mathcal{F})$ and $\hat{\mathcal{R}}_n(\mathcal{F})$ are the Rademacher average and empirical (conditional) Rademacher average of $\mathcal{F}$, respectively. Intuitive reason for  $\mathcal{R}_n(\mathcal{F})$ as a measure of complexity is that it quantifies the extent to which a function from the class $\mathcal{F}$ can correlate to random noise, a function belonging to a complex set can correlate to any random sequence. For a comprehensive overview of Rademacher averages and it's properties refer to \cite{mendelson2003few,bartlett2002rademacher,bartlett2005local}. Results from the aforementioned research work that have been used in the proofs related to our work are mentioned below. 

The following is the concentration inequality that depicts the relation between Rademacher averages and empirical Rademacher averages. The derivation utilizes Talagrand's inequality, kindly refer to Lemma A.4 in \cite{bartlett2005local} for full derivation.

\begin{lemma}
\label{lemma:rademacher_concentration}
   Let $\mathcal{F}$ be a class functions with range $\left[a,b\right]$. For fixed $\delta>0$, with probability of atleast $1-\delta$,
    \begin{equation*}
        \mathcal{R}_n(\mathcal{F}) \leq \inf\limits_{\alpha\in(0,1)}\left(
        \frac{1}{1-\alpha} \hat{\mathcal{R}}_n(\mathcal{F}) + \frac{(b-a)log(\frac{1}{\delta})}{4n\alpha(1-\alpha)}\right)
    \end{equation*}
\end{lemma}

The expected maximum deviation of empirical means from actual can be bounded by Rademacher averages as shown in the following bound. Check Lemman A.5 in \cite{bartlett2005local} for derivation.

\begin{lemma}
\label{lemma:rademacher_symmetric}
    For any class of function $\mathcal{F}$ we have,
    \begin{equation*}
        \max\left(\mathbb{E}\sup\limits_{f\in\mathcal{F}}\left(\mathbb{E}_{\mathbb{P}}\left[f\right]-\mathbb{E}_{\mathbb{P}_n}\left[f\right]\right),
        \mathbb{E}\sup\limits_{f\in\mathcal{F}}\left(\mathbb{E}_{\mathbb{P}_n}\left[f\right]-\mathbb{E}_{\mathbb{P}}\left[f\right]\right)\right) \leq 2\mathcal{R}_n(\mathcal{F})
    \end{equation*}
\end{lemma}

Where, $\mathbb{E}_{\mathbb{P}_n}\left[f\right]$ is the empirical mean given $n$ samples from $\mathbb{P}$ given by $\frac{1}{n}\sum_{i=1}^nf(X_i)$. Using Lemma \ref{lemma:rademacher_concentration} and Lemma \ref{lemma:rademacher_symmetric}, one can relate expected maximum deviation of empirical estimate from actual value to the empirical Rademacher averages. 

We would like to point a minor error in the derivation of the generalization error bound in \etal \cite{li2020automated} where Lemma \ref{lemma:rademacher_symmetric} has been used. In their work left hand side of the bound has been misinterpreted as maximum deviation instead of expected maximum deviation. To relate maximum deviation to Rademacher average we need another bound before Lemma \ref{lemma:rademacher_symmetric} which relates maximum deviation to expected maximum deviation. We will look at this corrected approach in the next section where we derive the generalization results for our work.  

The following simple structural result can be used to express Rademacher averages for a complex class of functions in terms of Rademacher averages of simple class of functions.
\begin{lemma}
    \label{lemma:rademacher_contraction}
    If $\phi:\mathbb{R}\rightarrow\mathbb{R}$ is Lipschitz with constant $L_\phi$ and satisfies $\phi(0) = 0$, then $\mathcal{R}_n(\phi\circ\mathcal{F}) \leq 2L_\phi\mathcal{R}_n(\mathcal{F})$ 
\end{lemma}

Next, we look at the empirical Rademacher average for the class of functions represented by our ASKL critic.

\begin{theorem}
        \label{lemma:askl_rademacher}
    The empirical Rademacher average of the RKHS $\mathcal{F}$ learned by the ASKL critic can be bounded and is described as follows, 
    \begin{align*}
        \hat{\mathcal{R}}_n(\mathcal{F}) \leq \frac{B}{n} \sqrt{\sum_{i=1}^n\lVert \phi\left(x_i\right)\rVert_2^2} \leq \frac{B}{\sqrt{n}}
    \end{align*}
    Where, $B = \sup\limits_{f\in\mathcal{F}}\lVert w \rVert_2$. 
\end{theorem}
\begin{proof}
    \begin{align}
        \hat{\mathcal{R}}_n(\mathcal{F}) &= \frac{1}{n} \mathbb{E}_{\sigma}\left[\sup\limits_{f \in \mathcal{F}}\left(\sum_{i=1}^n\sigma_i f\left(x_i\right)\right)\right] \nonumber \\
        &= \frac{1}{n} \mathbb{E}_{\sigma}\left[\sup\limits_{f \in \mathcal{F}}\left(w^\intercal\Phi_\sigma\right)\right]\nonumber
    \end{align}
    Here $\Phi_\sigma = \sum_{i=1}^n\sigma_i\phi(x_i)$ is a $D$ dimensional vector

    \begin{align}
        \hat{\mathcal{R}}_n(\mathcal{F}) &=\frac{1}{n} \mathbb{E}_{\sigma}\left[\sup\limits_{f \in \mathcal{F}}\left(w^\intercal\Phi_\sigma\right)\right] \nonumber \\
        &\leq \frac{1}{n}\mathbb{E}_{\sigma}\left[\sup\limits_{f \in \mathcal{F}}\left(\lVert w \rVert_2 \lVert \Phi_\sigma \rVert_2\right)\right] \label{step:rademacher_1}\\
        &\leq \frac{B}{n} \mathbb{E}_{\sigma}\left[\lVert \Phi_\sigma \rVert_2\right] \nonumber \\
        &\leq \frac{B}{n} \sqrt{\mathbb{E}_{\sigma}\left[\lVert \Phi_\sigma \rVert_2^2\right]} \label{step:rademacher_2}
    \end{align}
    Where step \ref{step:rademacher_1} is a direct implication of the Cauchy-Schwarz inequality.
    \begin{equation}
    \label{step:rademacher_3}
        \begin{gathered}            \mathbb{E}_{\sigma}\left[\lVert\Phi_\sigma\rVert_2^2\right] = \mathbb{E}_{\sigma}\left[\sum_{i=1}^n\sum_{j=1}^n\sigma_i\sigma_j\phi\left(x_i\right)^\intercal\phi\left(x_j\right)\right] = \sum_{i=1}^n\lVert \phi\left(x_i\right)\rVert_2^2 \\= \frac{1}{2D}\sum_{i=1}^n\sum_{j=1}^D cos\left(\left(\omega_j-\omega_j^\prime\right)^\intercal x_i\right) + 1 \leq n
        \end{gathered}
    \end{equation}

    From \ref{step:rademacher_2} and \ref{step:rademacher_3} we have the final result.
\end{proof}

\section{Generalization Error Bounds}
In this section we derive the generalization error bounds contributed in the scope of paper. We represent joint distribution, $\mathbb{P}_{XY}$ as  $\mathbb{P}$ and the product of marginal distributions, $\mathbb{P}_X\otimes\mathbb{P}_Y$, as $\mathbb{Q}$. Both distribution are define on measurable space $(\mathcal{X}\times\mathcal{Y},\Sigma_{XY})$. $\mathbb{P}_n$ and $\mathbb{Q}_m$ represents the corresponding empirical distributions and the pair $(x,y)$ is referred as $z$. The proofs use McDiarmid's inequality which is described as follows:

\begin{lemma}[McDiarmid's inequality]
    \label{lemma:mcdiarmids}
    Let $X_1,\dots,X_n$ be independent random variables taking values in a set $\mathcal{X}$, and assume that $\phi:\mathcal{X}^n\rightarrow\mathbb{R}$ satisfies
    \begin{equation*}
        \sup\limits_{x_1,\dots,x_n,x_i^{\prime}\in\mathcal{X}}\left|\phi(x_1,\dots,x_n) - \phi(x_1,\dots,x_{i-1},x_i^{\prime},x_{i+1},\dots,x_n)\right| \leq c_i
    \end{equation*}
    for every $1\leq i\leq n$. 
    
    Then, for every $t>0$,
    \begin{equation*}
        Pr\left\{\phi\left(X_1,\dots,X_n\right)-\mathbb{E}\left[\phi\left(X_1,\dots,X_n\right)\right]\geq t\right\} \leq e^{-2t^2/\sum_{i=1}^n c_i^2}
    \end{equation*}
    Stated in another way, for some fixed $\delta>0$ and with probability of at least $1-\delta$:

    \begin{equation*}
        \phi\left(X_1,\dots,X_n\right) \leq \mathbb{E} \left[\phi\left(X_1,\dots,X_n\right)\right] + \sqrt{\frac{\sum_{i=1}^n c_i^2 \; log\left(\frac{1}{\delta}\right)}{2}}
    \end{equation*}
\end{lemma}

In this section generalization error bounds for two lower bounds of mutual information $I_{TUBA}$ and $I_{DV}$ have been derived.
\begin{align}
    I_{TUBA}\left(f\right) &= \mathbb{E}_{\mathbb{P}}\left[f(z)\right] - \frac{\mathbb{E}_{\mathbb{Q}}\left[e^{f(z)}\right]}{a} - log\left(a\right) + 1 \label{eq:TUBA} \\
    I_{DV}\left(f\right) &= \mathbb{E}_{\mathbb{P}}\left[f(z)\right] - log\left(\mathbb{E}_{\mathbb{Q}}\left[e^{f(z)}\right]\right) \label{eq:DV}
\end{align}

Where in Eq.\ref{eq:TUBA} the baseline $a(y)$ is restricted to a constant $a$, this is because both $I_{MINE}$ and $I_{NWJ}$ lower bounds considered in this work correspond to constant baseline case. As the true distributions $\mathbb{P}$ and $\mathbb{Q}$ are unknown, we approximate the true expectation, with expectation with respect to empirical distributions $\mathbb{P}_n$ and $\mathbb{Q}_m$ corresponding to $n$ independent samples, $\left\{z_i\right\}_{i=1}^n$, from $\mathbb{P}$ and $m$ independent samples, $\left\{z_i^\prime\right\}_{i=1}^m$, from $\mathbb{Q}$ respectively.

\begin{align}
    \hat{I}_{TUBA}^{n,m}\left(f,S\right) &= \mathbb{E}_{\mathbb{P}_n}\left[f(z)\right] - \frac{\mathbb{E}_{\mathbb{Q}_m}\left[e^{f(z)}\right]}{a} - log\left(a\right) + 1 \label{eq:TUBA_empirical} \\
    \hat{I}_{DV}^{n,m}\left(f,S\right) &= \mathbb{E}_{\mathbb{P}_n}\left[f(z)\right] - log\left(\mathbb{E}_{\mathbb{Q}_m}\left[e^{f(z)}\right]\right) \label{eq:DV_empirical}
\end{align}
Where $S$ is the set of samples $\left\{z_i\right\}_{i=1}^n$ and $\left\{z_i^\prime\right\}_{i=1}^m$. The goal of generalization error bound is to bound the maximum deviation between $I_{TUBA}$ and $\hat{I}_{TUBA}$ or between $I_{DV}$ and $\hat{I}_{DV}$.

\begin{theorem}[Generalization bound for $I_{TUBA}$]
    \label{theorem:tuba_generalization}
    Assume that the function space $\mathcal{F}$ learnt by the critic is uniformly bounded by $M$, that is  $\left|f(z)\right| \leq M\; \forall f \in \mathcal{F}\;\&\; \forall z \in \mathcal{X}\times\mathcal{Y}$, where $M < \infty$. For any fixed $\delta >0$ generalization error of TUBA estimate can be bounded with probability of at least $1-\delta$

    \begin{equation*}
        \begin{split}
            \sup\limits_{f \in \mathcal{F}} \left(I_{TUBA}(f) - \hat{I}_{TUBA}^{n,m}(f)\right) &\leq 4 \hat{\mathcal{R}}_n\left(\mathcal{F}\right) + \frac{8}{a} e^M \hat{\mathcal{R}}_m\left(\mathcal{F}\right) + \frac{4M}{n}log\left(\frac{4}{\delta}\right) + \\
            &\frac{8Me^M}{am}log\left(\frac{4}{\delta}\right) + \sqrt{\frac{\left(\frac{4M^2}{n} + \frac{\left(e^M-e^{-M}\right)^2}{a^2m}\right) log\left(\frac{2}{\delta}\right)}{2}}
        \end{split}
    \end{equation*}
    Where, sample set $S$ for $\hat{I}_{TUBA}^{n,m}$ has been implicitly assumed to be given.
\end{theorem}
\begin{proof}
    Let,
    \begin{equation*}
        \phi(S) = \sup\limits_{f \in \mathcal{F}} \left(I_{TUBA}(f) - \hat{I}_{TUBA}^{n,m}(f,S)\right)
    \end{equation*}
    
    Letting $\Tilde{S}_i$ represent another set of samples which differ from $S$ at only one sample $z_i$ when $i\in\left[1,n\right]$ or at sample $z^\prime_i$ when $i\in\left[n+1,n+m\right]$, where the first case is when differing sample is sampled from $\mathbb{P}$ and the second case is when the differing sample is sampled from $\mathbb{Q}$. Now, when $i \in \left[1,n\right]$
    \begin{align}
        \left|\phi(S) - \phi(\Tilde{S}_i)\right| &= |\sup\limits_{f \in \mathcal{F}} \left(I_{TUBA}(f) - \hat{I}_{TUBA}^{n,m}(f,S)\right) \nonumber \\
        &\quad - \sup\limits_{f \in \mathcal{F}} \left(I_{TUBA}(f) - \hat{I}_{TUBA}^{n,m}(f,\Tilde{S}_i)\right)| \nonumber \\
        & \leq \sup \left|\hat{I}_{TUBA}^{n,m}(f,\Tilde{S}_i)-\hat{I}_{TUBA}^{n,m}(f,S)\right| \nonumber \\
        &= \frac{1}{n}sup \left|f(\Tilde{z}_i)-f(z_i)\right| \nonumber \\
        \left|\phi(S) - \phi(\Tilde{S}_i)\right| &\leq \frac{2M}{n} \label{step:tuba_1}
    \end{align}
    
    Where, step \ref{step:tuba_1} is because the maximum difference between values of a function bounded between $\left[-M,M\right]$ is $2M$, when  $i \in \left[n+1,n+m\right]$.
    
    \begin{flalign}
        \left|\phi(S) - \phi(\Tilde{S}_i)\right| &\leq \sup\limits_{f \in \mathcal{F}} \left|\hat{I}_{TUBA}^{n,m}(f,\Tilde{S}_i) - \hat{I}_{TUBA}^{n,m}(f,S)\right| \nonumber \\
        &= \frac{1}{am} \sup \left|e^{f\left(\Tilde{z}_i^\prime\right)} - e^{f\left(z_i^\prime\right)}\right| \nonumber \\
        \left|\phi(S) - \phi(\Tilde{S}_i)\right| &\leq \frac{e^M-e^{-M}}{am} \label{step:tuba_2}
    \end{flalign}
    
    and, step \ref{step:tuba_2} is due to the fact the maximum difference between values of exponential of a function bounded in $\left[-M,M\right]$ is $e^M - e^{-M}$. Because, there exists a $c_i$ such that $\left|\phi\left(S\right) - \phi\left(\Tilde{S}_i\right)\right| < c_i\; \forall\; i\in\left[1,m+n\right]$ we can apply McDiarmid's inequality, lemma \ref{lemma:mcdiarmids}, and write for a fixed $\delta > 0$ with probability of at least $1-\delta/2$, 
    \begin{equation}
    \label{step:tuba_mcdiarmids}
        \begin{gathered}
        \sup\limits_{f \in \mathcal{F}} \left(I_{TUBA}(f) - \hat{I}_{TUBA}^{n,m}(f)\right) \leq \mathbb{E}\left[\sup\limits_{f \in \mathcal{F}} \left(I_{TUBA}(f) - \hat{I}_{TUBA}^{n,m}(f)\right)\right] +\\
        \sqrt{\frac{\left(\frac{4M^2}{n} + \frac{\left(e^M-e^{-M}\right)^2}{a^2m}\right) log\left(\frac{2}{\delta}\right)}{2}}
        \end{gathered}
    \end{equation}
    
    By using lemma \ref{lemma:rademacher_symmetric} we get,
    
    \begin{equation}
        \label{step:tube_3}
        \mathbb{E} \left[\sup\limits_{f \in \mathcal{F}}\left(\mathbb{E}_{\mathbb{P}}\left[f\right] - \mathbb{E}_{\mathbb{P}_n}\left[f\right]\right)\right] \leq 2\mathcal{R}_n(\mathcal{F})
    \end{equation}
    
    Similarly, if we consider a family of functions $\psi\circ\mathcal{F} = \left\{\psi(f(z)): \forall f\in\mathcal{F}\right\}$ where $\psi(x) = e^x-1$.
    
    \begin{align}
        \mathbb{E}\left[\sup\limits_{f \in \mathcal{F}}\left(\mathbb{E}_{\mathbb{Q}_m}\left[e^f\right]-\mathbb{E}_{\mathbb{Q}}\left[e^f\right]\right)\right] &= \mathbb{E}\left[\sup\limits_{g \in \psi\circ\mathcal{F}}\left(\mathbb{E}_{\mathbb{Q}_m}\left[g\right]-\mathbb{E}_{\mathbb{Q}}\left[g\right]\right)\right] \nonumber \\
        &\leq 2\mathcal{R}_m\left(\psi\circ\mathcal{F}\right) \label{step:tuba_4}\\
    \mathbb{E}\left[\sup\limits_{f \in \mathcal{F}}\left(\mathbb{E}_{\mathbb{Q}_m}\left[e^f\right]-\mathbb{E}_{\mathbb{Q}}\left[e^f\right]\right)\right] &\leq 4e^M\mathcal{R}_m\left(\mathcal{F}\right) \label{step:tuba_5}
    \end{align}
    
    Step \ref{step:tuba_4} is from lemma \ref{lemma:rademacher_symmetric} and step \ref{step:tuba_5} is in implication of lemma \ref{lemma:rademacher_contraction} and the fact that $\phi(x) = e^x -1$ is Lipschitz with constant $e^M$ when $x \in \left[-M,M\right]$. Now, we are in a position to relate expectation of maximum deviation to Rademacher average.
    
    \begin{align}
        \mathbb{E}\left[\sup\limits_{f \in \mathcal{F}} \left(I_{TUBA}(f) - \hat{I}_{TUBA}^{n,m}(f)\right)\right] &\leq \mathbb{E} \left[\sup\limits_{f \in \mathcal{F}}\left(\mathbb{E}_{\mathbb{P}}\left[f\right] - \mathbb{E}_{\mathbb{P}_n}\left[f\right]\right)\right] + \nonumber \\
        &\quad\frac{1}{a}\mathbb{E}\left[\sup\limits_{f \in \mathcal{F}}\left(\mathbb{E}_{\mathbb{Q}_m}\left[e^f\right]-\mathbb{E}_{\mathbb{Q}}\left[e^f\right]\right)\right] \nonumber \\
        \mathbb{E}\left[\sup\limits_{f \in \mathcal{F}} \left(I_{TUBA}(f) - \hat{I}_{TUBA}^{n,m}(f)\right)\right] &\leq 2\mathcal{R}_n\left(\mathcal{F}\right) + \frac{4e^M}{a}\mathcal{R}_m\left(\mathcal{F}\right) \label{step:tuba_6}
    \end{align}
    
    The last step \ref{step:tuba_6} is in tandem with steps \ref{step:tuba_5} and \ref{step:tube_3}. Using lemma \ref{lemma:rademacher_concentration} and setting $\alpha = 1/2$ we get with probability of at least $1-\delta/4$:
    
    \begin{equation}
    \label{step:tuba_concentration}
        \mathcal{R}_n\left(\mathcal{F}\right) \leq 2\hat{\mathcal{R}}_n\left(\mathcal{F}\right) + \frac{2M log\left(\frac{4}{\delta}\right)}{n}
    \end{equation}
    
    Similar, relationship can also be stated between $\mathcal{R}_m\left(\mathcal{F}\right)$ and $\hat{\mathcal{R}}_m\left(\mathcal{F}\right)$
    
    Combining \ref{step:tuba_mcdiarmids}, \ref{step:tuba_6} and \ref{step:tuba_concentration} we get with probability of at least $1-\delta$
    
    \begin{align}
        \sup\limits_{f \in \mathcal{F}}\left(I_{TUBA}\left(f\right) - \hat{I}_{TUBA}^{n,m}\right) \leq& 4\hat{\mathcal{R}}_n\left(\mathcal{F}\right) + \frac{8e^M}{a}\hat{\mathcal{R}}_m\left(\mathcal{F}\right) + \frac{4M log\left(\frac{4}{\delta}\right)}{n} + \nonumber \\
        &\frac{8Me^M log\left(\frac{4}{\delta}\right)}{am} + \sqrt{\frac{\left(\frac{4M^2}{n} + \frac{\left(e^M-e^{-M}\right)^2}{a^2m}\right) log\left(\frac{2}{\delta}\right)}{2}}
    \end{align}
\end{proof}

Li \etal \cite{li2020automated} in derivation of generalization error bounds incorrectly replaced expectation of maximum deviation with maximum deviation in lemma \ref{lemma:rademacher_symmetric}. To rectify that error, we used McDiarmid's Inequality to bound maximum deviation with expected maximum deviation, this adds and additional term inside  square root in the bound in theorem \ref{theorem:tuba_generalization}.

Next, we are going to look at the generalization error bounds of Donsker Varadhan estimates, it is used to estimate mutual information in $I_{SMILE}$ estimate. We follow the same procedure used for deriving generalization error bounds of $I_{TUBA}$ to keep the proof brief. 

\begin{theorem}[Generalization bound for $I_{DV}$]
    Assume that the function space $\mathcal{F}$ learnt by the critic is uniformly bounded by $M$, that is  $\left|f(z)\right| \leq M\; \forall f \in \mathcal{F}\;\&\; \forall z \in \mathcal{X}\times\mathcal{Y}$, where $M < \infty$. For a fixed $\delta >0$ generalization error of Donsker Varadhan estimate can be bounded with probability of at least $1-\delta$
    
    \begin{equation*}
        \begin{split}
            \sup\limits_{f \in \mathcal{F}}\left(I_{DV}(f) - \hat{I}_{DV}^{n,m}(f)\right) \leq 4\hat{\mathcal{R}}_n\left(\mathcal{F}\right) + 8e^{2M}\hat{\mathcal{R}}_m\left(\mathcal{F}\right) + \frac{4M}{n} log\left(\frac{4}{\delta}\right)\\
            \frac{8Me^{2M}}{m} log\left(\frac{4}{\delta}\right) + \sqrt{\frac{\left(\frac{4M^2}{n} + \frac{\left(e^{2M}-1\right)^2}{m}\right) log\left(\frac{2}{\delta}\right)}{2}}
        \end{split}
    \end{equation*}
    Where, sample set $S$ for $\hat{I}_{DV}^{n,m}$ is implicitly assumed.
\end{theorem}

\begin{proof}
    Let
    \begin{equation*}
        \phi(S) = \sup\limits_{f \in \mathcal{F}} \left(I_{DV}(f) - \hat{I}_{DV}^{n,m}(f,S)\right)
    \end{equation*}
    Letting $\Tilde{S}_i$ represent another set of samples which differ from $S$ at only one sample. When $i \in \left[1,n\right]$ $\left|\phi(S) - \phi(\Tilde{S}_i)\right| \leq \frac{2M}{n}$ from equation \ref{step:tuba_1}, when $i \in \left[n+1,n+m\right]$
    
    \begin{align}
        \left|\phi(S) - \phi(\Tilde{S}_i)\right| &\leq \sup\limits_{f \in \mathcal{F}} \left|\hat{I}_{DV}\left(\Tilde{S}_i\right) - \hat{I}_{DV}\left(S_i\right)\right| \nonumber \\
        &= \sup \left|log\left(\mathbb{E}_{\mathbb{Q}_{m,i}}\left[e^{f\left(z\right)}\right]\right)-log\left(\mathbb{E}_{\mathbb{Q}_{m}}\left[e^{f\left(z\right)}\right]\right)\right| \label{step:dv_1}\\
        &\leq e^M \sup \left|\mathbb{E}_{\mathbb{Q}_{m,i}}\left[e^{f\left(z\right)}\right] - \mathbb{E}_{\mathbb{Q}_{m}}\left[e^{f\left(z\right)}\right]\right| \label{step:dv_2} \\
        &= \frac{e^M}{m} \sup \left|e^{f\left(\Tilde{z}_i^\prime\right)} - e^{f\left(z_i^\prime\right)}\right| \nonumber \\
        &\leq \frac{e^{2M}-1}{m} \label{step:dv_3}
    \end{align}
    In step \ref{step:dv_1} $\mathbb{Q}_{m,i}$ refers to the empirical distribution corresponding to the sample set $\Tilde{S}_i^\prime$. Where, inequality \ref{step:dv_2} is due to the fact that $log\left(x\right)$ is Lipschitz with constant $e^M$ when $x \in \left[e^{-M},e^{M}\right]$. Inequality \ref{step:dv_3} is due to the fact that the maximum difference between values of exponential of a function bounded in $\left[-M,M\right]$ is $e^M - e^{-M}$. Now, apply McDairmid's inequality to $\phi\left(S\right)$, for a fixed $\delta > 0$ with probability of at least $1-\delta/2$:
    
    \begin{equation}
    \label{step:dv_mcdiarmids}
        \begin{gathered}
        \sup\limits_{f \in \mathcal{F}} \left(I_{DV}(f) - \hat{I}_{DV}^{n,m}(f)\right) \leq \mathbb{E}\left[\sup\limits_{f \in \mathcal{F}} \left(I_{DV}(f) - \hat{I}_{DV}^{n,m}(f)\right)\right] +\\
        \sqrt{\frac{\left(\frac{4M^2}{n} + \frac{\left(e^{2M}-1\right)^2}{m}\right) log\left(\frac{2}{\delta}\right)}{2}}
        \end{gathered}
    \end{equation}
    
    We use \ref{step:tube_3} and \ref{step:tuba_5} to bound expected maximum deviation with Rademacher averages
    
    \begin{align}
        \mathbb{E}\left[\sup\limits_{f \in \mathcal{F}} \left(I_{DV}(f) - \hat{I}_{DV}^{n,m}(f)\right)\right] &\leq \mathbb{E} \left[\sup\limits_{f \in \mathcal{F}}\left(\mathbb{E}_{\mathbb{P}}\left[f\right] - \mathbb{E}_{\mathbb{P}_n}\left[f\right]\right)\right] + \nonumber \\
        &\quad \mathbb{E} \left[\sup\limits_{f \in \mathcal{F}}\left(log\left(\mathbb{E}_{\mathbb{Q}_m}\left[e^f\right]\right) - log\left(\mathbb{E}_{\mathbb{Q}_m}\left[e^f\right]\right)\right)\right] \nonumber \\
        &\leq \mathbb{E} \left[\sup\limits_{f \in \mathcal{F}}\left(\mathbb{E}_{\mathbb{P}}\left[f\right] - \mathbb{E}_{\mathbb{P}_n}\left[f\right]\right)\right] + \nonumber \\
        &\quad e^M \mathbb{E} \left[\sup\limits_{f \in \mathcal{F}}\left(\mathbb{E}_{\mathbb{Q}_m}\left[e^f\right] - \mathbb{E}_{\mathbb{Q}_m}\left[e^f\right]\right)\right] \nonumber \\
        \mathbb{E}\left[\sup\limits_{f \in \mathcal{F}} \left(I_{DV}(f) - \hat{I}_{DV}^{n,m}(f)\right)\right] &\leq 2\mathcal{R}_n\left(\mathcal{F}\right) + 4e^{2M}\mathcal{R}_m\left(\mathcal{F}\right) \label{step:dv_4}
    \end{align}
    Where last step is in tandem with \ref{step:tube_3} and \ref{step:tuba_5}. Combining \ref{step:dv_mcdiarmids}, \ref{step:dv_4} with \ref{step:tuba_concentration} we get the final result with probability of at least $1-\delta$ described below as,
    \begin{equation}
        \begin{split}
            \sup\limits_{f \in \mathcal{F}}\left(I_{DV}(f) - \hat{I}_{DV}^{n,m}(f)\right) \leq 4\hat{\mathcal{R}}_n\left(\mathcal{F}\right) + 8e^{2M}\hat{\mathcal{R}}_m\left(\mathcal{F}\right) + \frac{4M}{n} log\left(\frac{4}{\delta}\right)\\
            \frac{8Me^{2M}}{m} log\left(\frac{4}{\delta}\right) + \sqrt{\frac{\left(\frac{4M^2}{n} + \frac{\left(e^{2M}-1\right)^2}{m}\right) log\left(\frac{2}{\delta}\right)}{2}}
        \end{split}
    \end{equation}
\end{proof}
 
{\bibliographystyle{IEEEtran}
\bibliography{supplement}
}